\documentclass[preprint]{revtex4}

\usepackage{amsfonts,xfrac}
\usepackage{amsmath}
\usepackage{amssymb}
\usepackage{graphicx}
\usepackage{dcolumn}
\usepackage{dsfont}
\usepackage{times}
\usepackage{epsfig,color}
\usepackage[]{units}
\usepackage{soul}
\usepackage{bm}
\usepackage[hang,nooneline]{subfigure}                         
\newcounter{fig}   

\def\epsilon{\varepsilon}

\begin{document}

\title{\textsf{Leaf-like Origami with Bistability for Self-Adaptive Grasping Motions}}
\author{Hiromi Yasuda$^{1, 2}$, Kyle Johnson$^{3}$\footnote{First two authors contributed equally to this work.}, Vicente Arroyos$^{3}$, Koshiro Yamaguchi$^{1}$, Jordan R. Raney$^{2}$, and Jinkyu Yang$^{1}$}
\affiliation{$^1$Department of Aeronautics \& Astronautics, University of Washington, Seattle, WA 98195-2400, USA\\ $^2$Department of Mechanical Engineering and Applied Mechanics, University of Pennsylvania, Philadelphia, PA 19104, USA\\ $^3$Department of Electrical and Computer Engineering, University of Washington, Seattle, WA 98195-2400, USA}

\begin{abstract}
The leaf-like origami structure was inspired by geometric patterns found in nature, exhibiting unique transitions between open and closed shapes. With a bistable energy landscape, leaf-like origami is able to replicate the autonomous grasping of objects observed in biological systems like the Venus flytrap. 
We show uniform grasping motions of the leaf-like origami, as well as various non-uniform grasping motions which arise from its multi-transformable nature. Grasping motions can be triggered with high tunability due to the structure's bistable energy landscape.
We demonstrate the self-adaptive grasping motion by dropping a target object onto our paper prototype, which does not require an external power source to retain the capture of the object.
We also explore the non-uniform grasping motions of the leaf-like structure by selectively controlling the creases, which reveals various unique grasping configurations that can be exploited for versatile, autonomous, and self-adaptive robotic operations. 
\end{abstract}

\maketitle

\clearpage
\section*{Introduction}
Plants and animals are perpetually subjected to the unpredictable nature of their ever-changing surroundings, forcing the development of dynamic responses and self-adaptive behavior to accommodate for their rapidly changing environment. For example, sea anemones can catch targets with different shapes and sizes because of their self-adaptive morphology~\cite{Zang2020}, and the Venus flytrap is capable of self-adaptive movements in response to a multitude of stimuli~\cite{Guo2015,Mu2018}.
Engineering a versatile grasping device that can mimic 
the nastic movements of these natural adaptive systems
is one of the many challenges facing the field of robotics. 

Recently, origami-inspired structures have been studied with a focus on biomimetic folding/unfolding behavior. 
For example, one of the most well-known crease patterns, the Miura-ori~\cite{Miura1985}, can be found in tree leaves and other plants~\cite{Kobayashi1998,Focatiis2002,Mahadevan2005}.
Additionally, a fast folding grasping motion inspired by earwig wings has been demonstrated by introducing an extensional membrane element to the Miura-ori building block~\cite{Faber2018}.
These Miura-ori based reconfigurable structures show great potential for engineering devices, but the variations of their folding patterns are still limited to the single degree of freedom (DOF) 
nature of the Miura-ori fold implemented using a rigid origami approach. These limitations have hindered the adoption of Miura-ori 
in practical applications.


Using the Miura-ori unit cell as a building block, De Focatiis and Guest explored the biomimetic leaf-like origami patterns, specifically leaf-in and leaf-out origami~\cite{Focatiis2002} (see Fig.~\ref{fig:Geometry} for the leaf-out origami).
It has been shown that leaf-out origami can exhibit numerous rigid folding motions (i.e., multi-DOF) despite the single DOF nature of the individual Miura-ori elements~\cite{Yasuda2016}.
Another interesting aspect of the leaf-out origami structure is its tunable bistability.
Despite such versatile 
control of the DOF and stability characteristics, leaf-out origami has not been explored much in engineering applications.

Here, we demonstrate the feasibility of 
using leaf-out origami's reconfigurability and bistability to achieve versatile, autonomous, and self-adaptive grasping motions.
First, we employ a rigid origami model to characterize the unique kinematics of the structure, especially transitions between open and closed modes separated by the flat state.
In addition, we model the crease folding as a torsional spring (e.g., waterbomb origami~\cite{Hanna2014,Treml2018}, degree-four vertex origami~\cite{Waitukaitis2015}, etc.), and we examine the tailorable energy landscapes by altering the design parameters, specifically the rest angle of the spring element.
We then design and fabricate multi-layered prototypes of the leaf-out origami structure with tailored energy barriers. 
We experimentally demonstrate the feasibility of catching a falling object by triggering the grasping motion, which only occurs within a predetermined range of impact forces.
Further, we numerically explore various different grasping configurations by controlling selected Miura-ori unit cells. 
Our numerical analysis reveals the multi-transformable morphology of the leaf-out origami, such as pinching folds between two tips of the structure and a symmetric grasping motion similar to that of alligator-clips.
Based on the controllable dynamic response of the origami-based structure, our design principles demonstrate a great potential for manipulating reconfigurable systems for various engineering applications such as surgical robots, space deployable structures, and catch-and-release traps. 

%
%

\section*{Leaf-out origami model}
\begin{figure}[htbp]
\centerline{ \includegraphics[width=1.\textwidth]{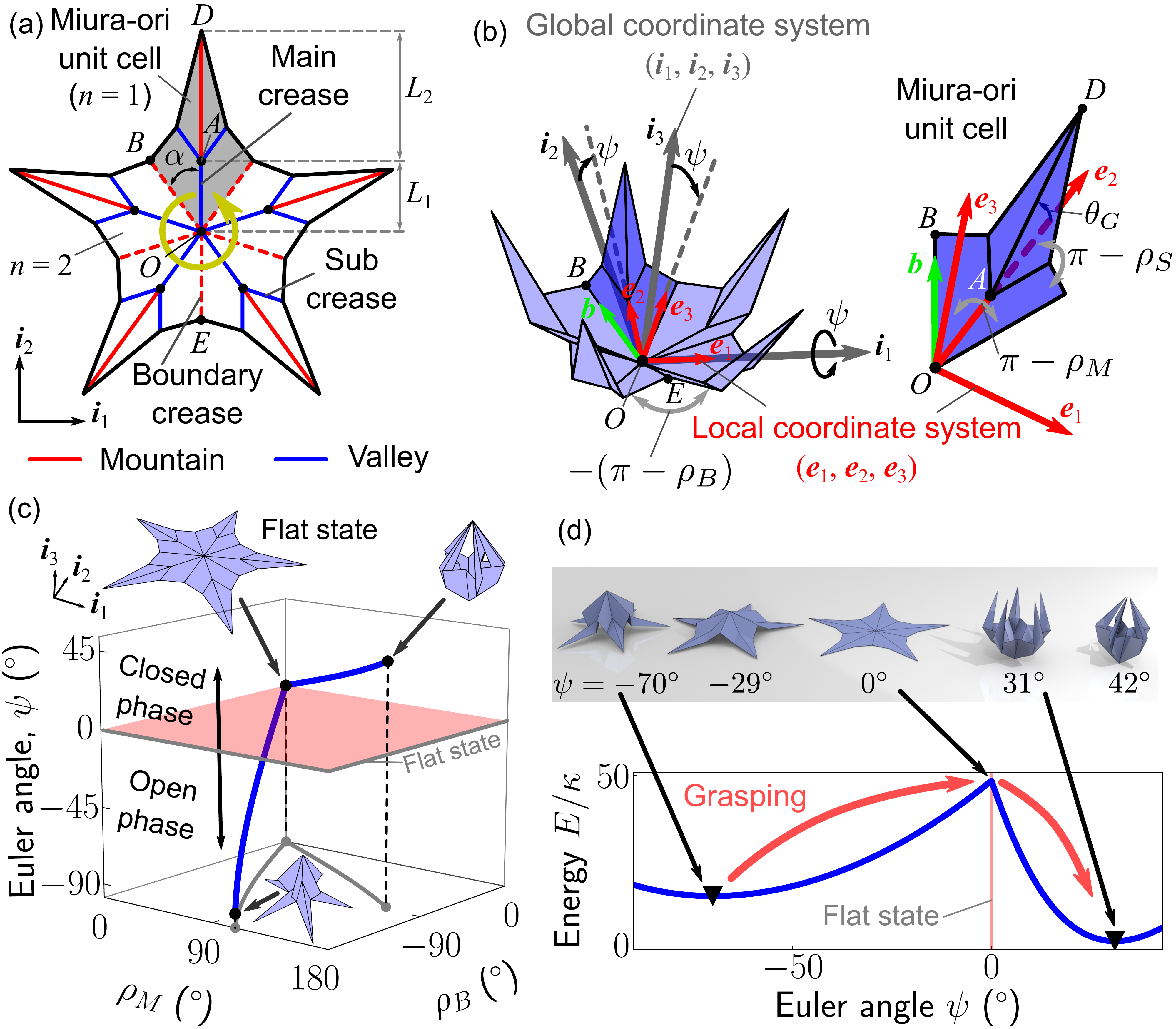}}
\caption{Leaf-out origami for grasping objects based on bistability. (a) Flat configuration of the leaf-out origami composed of five Miura-ori unit cells. Red (blue) lines indicate mountain (valley) crease lines. Dashed lines indicate the mountain boundary crease lines connecting adjacent Miura-ori unit cells. (b) Folded configuration of the leaf-out origami. We define the global coordinate as well as the local coordinate attached to one of the Miura-ori unit cell. The folding angle along the boundary crease is defined as $\rho_B$. Right inset illustrates the geometry of the Miura-ori unit. We define two folding angles $\rho_M$ and $\rho_S$ for the main and sub crease lines, respectively. (c) The uniform grasping motion of the leaf-out origami is plotted in the configuration space. The folding path projected onto the bottom plane ($\rho_M \rho_B$-plane) is denoted by the grey line. (d) The leaf-out origami can exhibit bistable behavior via transformation from its folded shape (open phase) into another folded configuration (closed phase) passing through the flat state. $( \bar{\rho}_M, \bar{\rho}_B) = ( 120^\circ, -30^\circ)$ is used for this calculation. }
\label{fig:Geometry}
\end{figure}

The leaf-out origami consists of $n_{cell}$ Miura-ori unit cells 
(see Fig.~\ref{fig:Geometry}(a)), with neighboring unit cells connected by the boundary crease lines~\cite{Focatiis2002,Yasuda2016}. 
The Miura-ori is known as a rigid origami structure whose surfaces can be modeled as rigid panels connected by hinges, and exhibits 1 degree-of-freedom (DOF) folding motion.
The geometry of each Miura-ori unit cell is characterized by the following geometrical parameters: central angle ($\alpha=\pi/n_{cell}$) and length parameters ($L_1=\overline{OA}$ and $L_2=\overline{AD}$).
Based on the 1 DOF nature of the Miura-ori unit cell, the folding motion of the whole leaf-out origami can be expressed by considering the orientation and folding motion of each Miura-ori unit cell.

To describe the orientation of the unit cell, we introduce the global coordinates $(\boldsymbol{i}_1, \boldsymbol{i}_2, \boldsymbol{i}_3)$ and the local coordinates $(\boldsymbol{e}_1, \boldsymbol{e}_2, \boldsymbol{e}_3)$ attached to the Miura-ori unit cell such that the $\boldsymbol{e}_2$-axis is aligned with $\overline{OA}$ and the direction of $\boldsymbol{e}_3$ is the same as the $\boldsymbol{i}_3$-axis at the flat state (see Fig.~\ref{fig:Geometry}(b)).
By using these coordinate systems, we define the Euler angle ($\psi$) to express the rotation of the local coordinate system around the $\boldsymbol{i}_1$-axis.
Here, if the leaf-out origami shows uniform grasping motion the negative (positive) Euler angle $\psi$ indicates the open (closed) state as shown in Fig.~\ref{fig:Geometry}(c).
We will later discuss the details of this uniform grasping motion.

Let $n$ be the unit index numbered counterclockwise. The folding motion of $n$th Miura-ori unit cell is characterized by the two folding angles, $\rho_{M,n}$ along the main crease lines and $\rho_{S,n}$ along the sub crease lines as shown in Fig.~\ref{fig:Geometry}(b).
Note that there are relationships between $\rho_{M,n}$ and $\rho_{S,n}$ (see Supplementary Note 1), therefore by knowing one of these angles we can determine the configuration of the unit cell structure.
The folding angle for the boundary crease, which connects the $n$th and the $(n+1)$th unit cells, is denoted by $\rho_{B,n}$.
Here, the folding angle ($\rho_j$) along the crease line $j$ is defined as the complement of the angle between adjacent surfaces connected by the crease $j$, so that positive (negative) folding angles indicate valley (mountain) folds.
In this study, we assume that the leaf-out origami maintains its mountain and valley crease assignments during folding/unfolding motions, i.e., $0\le \rho_M, \rho_S \le \pi$ and $-\pi \le \rho_B \le 0$.
 
To analyze the unique multi-transformable feature of the leaf-out structure~\cite{Yasuda2016} we employ the rigid origami simulation technique based on~\cite{Tachi2009,Tachi2010} and explore various grasping motions.
First, we consider the loop closure constraint around vertex $O$ which is connected to $N=2n_{cell}$ crease lines including main and boundary creases (see the yellow arrow in Fig.~\ref{fig:Geometry}(a)) as follows:
 \begin{equation} \label{eq:Loop}
	\bm{F}({\bm{\rho }_{O}})={{\bm{\chi}}_{1}}{{\bm{\chi}}_{2}}\cdots {{\bm{\chi}}_{N}}=\bm{I}_3,
\end{equation}
where $\bm{I}_{m}$ is the $m \times m$ identity matrix, and $\bm{\rho}_O=\left[  \rho_{M,1}  \ \  \rho_{B,1}  \ \   \cdots  \ \   \rho_{M,n_{cell}}   \ \   \rho_{B,n_{cell}}  \right]^T$ is the vector composed of $N$ folding angles along crease lines connected to vertex $O$.
Also, $\bm{\chi}_j$ represents the rotation matrix for the crease $j$ as a function of the folding angle $\rho_j$ and the central angle $\alpha$ (see Supplementary Note 2 for the details).
Then, we seek a solution of increment folding angles $\Delta \bm{\rho} = \left[  \Delta\rho_{M,1} \ \  \Delta\rho_{B,1}  \ \  \cdots  \ \  \Delta\rho_{M,n_{cell}}  \ \  \Delta\rho_{B,n_{cell}}  \right]^T$ such that the residual elements of $\bm{F}$, expressed by $\bm{r}  = \left[  {{r}_{a}}  \ \ {{r}_{b}} \ \  {{r}_{c}}  \right]^T$, vanishes.
Here, the residual components are calculated from
\begin{equation} \label{eq:rotation}
\bm{F} \left( \bm{\rho}^{(0)}+\Delta \bm{\rho} \right) = \left[ \begin{matrix}
   1 & 0-{{r}_{a}} & 0+{{r}_{c}}  \\
   0+{{r}_{a}} & 1 & 0-{{r}_{b}}  \\
   0-{{r}_{c}} & 0+{{r}_{b}} & 1
\end{matrix} \right],
\end{equation}
where the superscript $(0)$ indicates the initial values of the folding angles for the numerical analysis.
By introducing Euler integration~\cite{Tachi2009,Tachi2010} we calculate this increment $\Delta \bm{\rho}$ with respect to the valid initial state as follows:
\begin{equation} \label{eq:rotation}
	\Delta \bm{\rho} =\left[ {{\bm{I}}_{N}}-{{\bm{C}}^{+}}\bm{C} \right]\Delta {\bm{\rho }_{0}}-{{\bm{C}}^{+}} \bm{\xi},
\end{equation}
where $\bm{C}$ is the global linearized constraint matrix, and $\bm{C}^{+}$ is the Moore-Penrose pseudo-inverse matrix of $\bm{C}$.
Also, $\Delta \bm{\rho}_0$ denotes an arbitrary vector, and this vector is projected into the constrained subspace via $\left[ {{\bm{I}}_{N}}-{{\bm{C}}^{+}}\bm{C} \right]$ to find the valid increment of the folding angle closest to $\Delta \bm{\rho}_0$.
To compensate for the accumulation of numerical error due to the Euler integration method, we introduce the $3 \times 1$ vector $\bm{\xi}$ which is numerically obtained by solving $\bm{r}\left( \bm{\rho}^{(0)}+\Delta \bm{\rho}, \bm{\xi} \right)=\bm{0}$.
Therefore, we manipulate $\Delta \bm{\rho}_0$ by controlling some of the folding angles, and we simulate the folding motion of the leaf-out origami by using this approach (see Supplementary Note 2 for how we embed controlled angles in this approach).
 
Based on the kinematic analysis discussed above, we conduct the energy analysis by modeling crease line $j$ of the leaf-out origami as a linear torsion spring with spring constant $\kappa_j$.
By using this torsion spring model, we analyze the energy landscape as follows:
\begin{equation} \label{eq:energy}
	E =\frac{1}{2}\sum\limits_{j=1}^{{{N}_{Total}}}{{{\kappa }_{j}}{{\left( \rho_j -{{{\bar{\rho }}}_{j}} \right)}^{2}}},
\end{equation}
where $N_{Total}$ is the total number of crease lines in the leaf-out origami, and $\bar{\rho}_j$ is the rest folding angle of $j$th crease line.

 \section*{Uniform grasping motion}
In this section, we consider the uniform deployment/contraction, i.e., all of the Miura-ori unit cells show identical folding motion ($\rho_{M,n}=\rho_M$ and $\rho_{B,n}=\rho_B$ for $n=1, \dots, n_{cell}$), which corresponds to the case where the leaf-out origami catches a spherical target object.
To achieve this folding mode, we control $\Delta \bm{\rho}_0$ depending on the posture of each Miura-ori unit cell, which is described by the Euler angle ($\psi$).
Based on~\cite{Yasuda2016}, we can obtain the folding angle along the main crease line $\rho_M$ from the Euler angle $\psi$ as follows:
\begin{equation} \label{eq:uni_cond}
\tan \alpha =\frac{\sin \alpha \cos (\rho_M/2)} {\cos \alpha \cos \psi -\sin \alpha \sin \psi \sin (\rho_M/2)  }.
\end{equation}

Also, we examine the relationship between the  folding angle $\rho_B$ and the Euler angle $\psi$.
We consider the unit vector $\bm{b} = \left[ b_1 \ \ b_2 \ \ b_3 \right]^T$ along the boundary crease $\overline{OB}$ as shown in Fig.~\ref{fig:Geometry}(b), which is expressed by
 \begin{equation} \label{eq:b_vector}
\bm{b}  =\left[ {{\bm{i}}_{1}}  \ \ {{\bm{i}}_{2}} \ \ {{\bm{i}}_{3}}   \right]
\left[ \begin{matrix}
  -\sin \alpha \cos (\rho_M/2) \\ 
  \cos \alpha \cos \psi -\sin \alpha \sin \psi  \sin (\rho_M/2)  \\ 
  \cos\alpha \sin\psi +\sin \alpha \cos \psi  \sin (\rho_M/2)  \\ 
\end{matrix} \right].
\end{equation}
For the uniform grasping motion, the orientation and folding motion of all the Miura-ori unit cells are identical, which means that the rotation of the unit vector $\bm{e}_2$ around the boundary crease $\bm{b}$ through the angle $\rho_B$ is identical to the rotation of $\bm{e}_2$ around $\bm{i}_3$ through $2\alpha$.
The rotation around the vector $\bm{b}$ can be expressed by using the Rodrigues rotation formula as follows:
 \begin{equation} \label{eq:Rodrigues}
 {{\bm{R}}_{\bm{b}}}={{\bm{I}}_{3}}+\sin \left( -\pi +{{\rho }_{B}} \right)\bm{\hat{b}}+\left[ 1-\cos \left( -\pi +{{\rho }_{B}} \right) \right]{{\bm{\hat{b}}}^{2}}
\end{equation}
where
\begin{equation} \label{eq:b_mat}
\bm{\hat{b}}=\left[ \begin{matrix}
   0 & -{{b}_{3}} & {{b}_{2}}  \\
   {{b}_{3}} & 0 & -{{b}_{1}}  \\
   -{{b}_{2}} & {{b}_{1}} & 0  \\
\end{matrix} \right].
\end{equation}
Also, the rotation around $\bm{i}_3$ is expressed by
 \begin{equation} \label{eq:R3_matrix}
 {{\mathbf{R}}_{3}}=\left[ \begin{matrix}
   \cos 2\alpha  & -\sin 2\alpha  & 0  \\
   \sin 2\alpha  & \cos 2\alpha  & 0  \\
   0 & 0 & 1  \\
\end{matrix} \right].
\end{equation}
Therefore, we obtain
 \begin{equation} \label{eq:R3_matrix}
 {{\bm{R}}_{3}}{{\bm{e}}_{2}}={{\bm{R}}_{\bm{b}}}{{\bm{e}}_{2}}
\end{equation}
By solving this equation numerically, together with Eq.~\eqref{eq:uni_cond}, we obtain the relationship between the folding angles ($\rho_M$ and $\rho_B$), and the Euler angle $\psi$.

By feeding these folding angles $\rho_M$ and $\rho_B$ as a function of the Euler angle $\psi$ into the rigid origami model, we perform the simulation for the grasping motion of the leaf-out origami with $n_{cell}=5$.
Figure~\ref{fig:Geometry}(c) shows the folding path plotted in the 3D configuration space where the negative and positive Euler angle regimes correspond to the open and closed phases, respectively, and $\psi=0$ indicates that the leaf-out is completely flat (see the inset illustrations in the figure).

Based on the kinematic analysis, we also analyze the energy landscape for the transformation from the open phase to the closed state by using the energy expression (Eq.~\eqref{eq:energy}).
Assuming that the identical torsional stiffness for every crease line (i.e., $\kappa_j=\kappa$ for $j=1,\dots,N_{Total}$), we calculate the potential energy change as shown in Fig.~\ref{fig:Geometry}(d).
Here, the rest angles for the main and boundary crease liens are $( \bar{\rho}_M, \bar{\rho}_B) = ( 120^\circ, -30^\circ)$ for this calculation.
In the figure, the two black triangles indicate the local minimum states which are separated by the energy peak at the flat state ($\psi=0^\circ$).
Therefore, starting from the open phase, once the structure overcomes the energy barrier it triggers the grasping motion transforming to the stable closed state. To maintain this grasping shape, our leaf-out origami does not require external energy input, unlike soft robotic grippers with pneumatic actuation~\cite{Truby2018} and origami-based robotic grippers~\cite{Edmondson2013,Mintchev2018,Li2019,Hu2020}.

\begin{figure}[htbp]
\centerline{ \includegraphics[width=1.\textwidth]{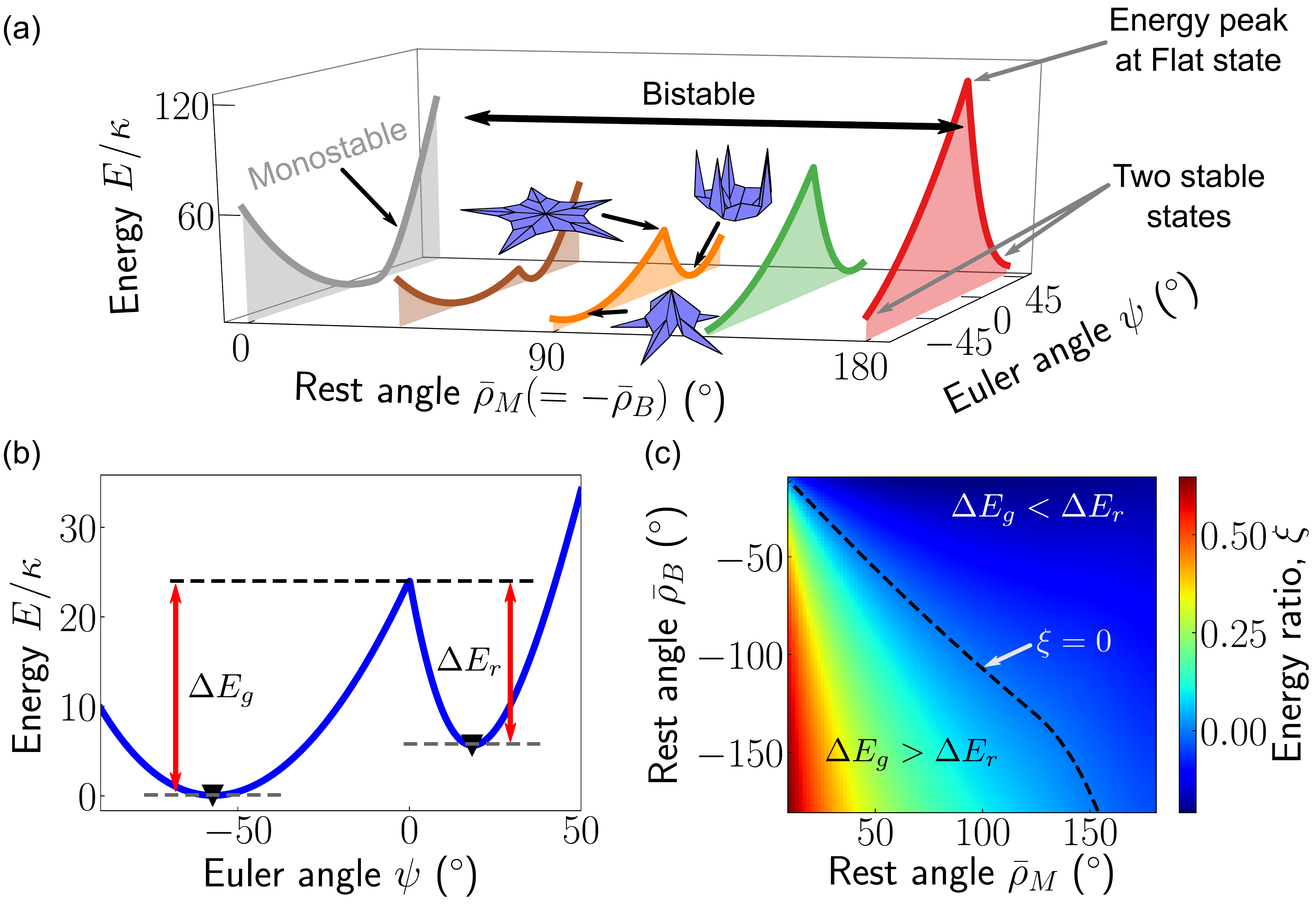}}
\caption{Energy analysis. (a) Tailorable energy landscape altered by the rest angle $\bar{\rho}_M$. 
Here, we assume $\bar{\rho}_M=-\bar{\rho}_B$, and $\bar{\rho}_S$ is obtained 
from the folding angle relationships (see Supplementary Note 1).
(b) We characterize the asymmetric energy landscape by defining the energy gaps, $\Delta E_g$ and $\Delta E_r$. $\left( \bar{\rho}_M, \bar{\rho}_B \right) = (60^\circ, -120^\circ)$ is used to calculate the potential energy curve. (c) Surface plot of the energy ratio, defined as $\xi=(\Delta E_g - \Delta E_r)/(\Delta E_g + \Delta E_r)$, as a function of two different rest angles $\left( \bar{\rho}_M, \bar{\rho}_B \right)$.}
\label{fig:Energy}
\end{figure}

The natural follow-up question: is the grasping motion tunable or not? In other words, can we tailor the energy landscape to control the energy barrier and locations of the energy minima?
To investigate the tunability of the grasping motion, we analyze the energy landscape for various configurations, specifically different rest angles.
For the sake of simplicity, we assume $\bar{\rho}_M=-\bar{\rho}_B$, and then $\bar{\rho}_S$ is obtained from the folding angle relationships (see Supplementary Note 1).
Also, we assume identical torsional stiffness ($\kappa$) for every crease line.
Figure~\ref{fig:Energy}(a) shows the energy landscapes as a function of the rest angle ($\bar{\rho}_M$) and Euler angle ($\psi$).
The energy curves of the leaf-out origami show the local maximum peak at $\psi=0^\circ$, which corresponds to the complete flat state, and the analysis result clearly shows two energetically lower states before/after this peak, i.e., bistable behavior.
In addition, this bistable energy curves can be tailored drastically by varying the rest angle ($\bar{\rho}_M$). 
Leaf-out origami can exhibit not only tunable bistability, but also a monostable configuration, specifically if $\bar{\rho}_M=0^\circ$, which corresponds to the case in which the leaf-out origami is initially in the flat state (see the grey colored curve in Fig.~\ref{fig:Energy}(a)).

To further analyze the tunability of this origami structure, we characterize the energy barrier between two stable states, by defining the energy gaps ($\Delta E_g$ and $\Delta E_r$) as shown in Fig.~\ref{fig:Energy}(b).
Letting the energy ratio be $\xi=(\Delta E_g - \Delta E_r)/(\Delta E_g + \Delta E_r)$, we compare the energy levels at each stable state by controlling two rest angles, $\bar{\rho}_M$ and $\bar{\rho}_B$.
The surface plot of the energy ratio ($\xi$) 
in Fig.~\ref{fig:Energy}(c) shows the two different regions bounded by the dashed line indicated $\xi=0$, i.e., the open and closed stable states show the exact same values of local minimum energy.
In the left lower region, $\Delta E_g$ is greater than $\Delta E_r$ (e.g., Fig.~\ref{fig:Energy}(b)), which means that if the origami starts from the open stable state and transits to the other, the energy difference $\Delta E_g - \Delta E_r$ is trapped in the system, which can be advantageous for 
applications involving impact.
On the other hand, if $\Delta E_g < \Delta E_r$, less energy is required to trigger the transition from open to closed states, a feature that can be particularly useful for sensitive actuation.

\section*{Experimental demonstration}
Based on the 
bistability discussed in the previous section, we fabricate a physical prototype to demonstrate the feasibility of a leaf-out origami grasper.
We design a multi-layered structure to enhance the rigidity of each facet as well as the folding behavior of the crease lines, which achieves rigid origami-like folding motions, i.e., deformation only takes place along the crease lines.
Figure~\ref{fig:Fabrication}(a) shows the layout of our leaf-out origami prototype composed of 6 layers stacked in the vertical direction.
We use construction paper sheets (thickness of 
0.5 mm) for the flat surfaces, and nylon (thickness of 0.025 mm) and polyester plastic (PET; thickness of 0.25 mm) sheets to obtain robust folding of the creases.
We cut each layer by using a laser cutting machine, and to bond each layer we insert adhesive layers (see the Methods section for details).
Figure~\ref{fig:Fabrication}(b) shows the fabricated prototype whose design parameters are ($n_{cell}$, $L_1$, $L_2$) = (5, 70~mm, 30~mm).
To achieve sharp and straight crease folding we introduce a comb-like crease design (see inset, Fig.~\ref{fig:Fabrication}(b)), which also helps to improve the consistency of folding behavior and to enhance the repeatability of tests.
Without the PET layer, the structure's crease line stiffness is defined primarily by the flexible nylon layer.
To increase the bending rigidity, the PET layer is applied and the crease line stiffness is assumed to be defined solely by the PET layer due to the pliability of nylon.

We examine the bistable energy landscape of our prototype by approximating the folding behavior of crease lines by a torsion spring.
To analyze and enhance the repeatable folding/unfolding motion of the creases, we perform cyclic loading tests (100 cycles) on single-crease samples (see Supplementary Fig.~1(b)).
We use the loading curve from the last (100th) cycle to approximate the torsion spring constant per unit crease width ($\kappa_{PET}=0.76$ Nm/rad/mm) and the rest angle $\bar{\rho}=71.8^\circ$ based on the rigid crease model~\cite{Yasuda2019tmp}. 
For the crease without PET, we only use the thin nylon sheet as an elastic hinge, so the force is extremely small.
Therefore, we neglect the effect of the creases without PET in the following analysis (i.e., $\kappa_M=\kappa_S=0$).
Finally, we use Eq.~\eqref{eq:energy} and construct the energy landscape of our actual prototype as shown in Fig.~\ref{fig:DropTest}(a).
The analysis result clearly indicates the two local minima in the potential energy plot, and each stable configuration of the leaf-out origami agrees qualitatively well with the postures of the actual prototype at its corresponding stable state.

To further validate the grasping based on bistability, specifically the energy barrier predicted by the rigid origami model, we conduct a drop test in which a polyurethane ball of mass $m_{ball}=22.3$~g and radius $R_{ball}=3.5$~cm is used as a target object.
The ball is pinched in place over the center of the leaf-out structure by pressure from a screw, and we simply unscrew the bolt to release the ball with zero initial velocity (see Supplementary Fig.~2 for the setup; see also the Methods section for details).
To capture the grasping motion, we use a high-speed camera (Chronos, Kron Technologies) filming at 1052 frames per second. 
Figure~\ref{fig:DropTest}(b) shows the snapshots from the drop test in which the drop height is 360~mm.
Our leaf-out origami prototype successfully demonstrates uniform grasping motion to capture the target ball (see also Supplementary Movie S1 for the grasping motion of the prototype). 
It continues holding the ball 
even after the snap-through behavior is triggered.

We analyze how the grasping motion can be triggered depending on the input, specifically the impact momentum, by varying the drop height.
If the ball is dropped at the height $h$, the potential energy at the initial state is calculated as $E_{ball}=m_{ball}gh$ where $g$ is the gravitational acceleration.
Let $E_{Gap}=(E_{ball}-\Delta E_g)/\kappa_{PET}$. We plot this energy difference as a function of the drop height ($h$) and rest angle ($\bar{\rho}$). This is shown in Fig.~\ref{fig:DropTest}(c), in which the dashed line indicates $E_{Gap}=0$.
If the drop height is below the threshold value ($E_{Gap}$) the grasping motion is not triggered (denoted by the cross markers). However, by crossing this boundary the leaf-out starts to overcome the energy barrier and transform into the 
closed phase
(denoted by the circle markers).
Note that if the impact momentum is too large, although the grasping motion is triggered, the leaf-out fails to retain the ball inside the structure (see Supplementary Movie S1 for this case).
Although our paper prototype 
demonstrates the feasibility of grasping motion within a limited impact range, 
it suggests the potential of adaptive grasping via sensing the velocity and location of impact 
based on the carefully designed energy landscape (see Supplementary Movie S2 for off-center grasping).

\section*{Exploring multi-grasping motions}
One of the important tasks for grasping applications is to catch/hold an object even if its target shape varies from object to object.
By leveraging the multi-DOF feature of the leaf-out origami, we show that the leaf-out can exhibit various grasping shapes.
To explore different folding motions, we selectively control the main crease line of Miura-ori unit cells.
In particular, we apply the controlled angle ($\Delta \rho_C$) to the main folding angle of $n$th Miura-ori unit cell ($\rho_{M,n}$) in the arbitrary vector $\Delta \bm{\rho}_0$, and then we perform the simulation to obtain the other main and boundary folding angles.
Since we are interested in different grasping shapes, we focus on the closed phase of the leaf-out with $n_{cell}=5$ in our numerical analysis.
We start the simulations from the slightly folded configuration, specifically $(\rho_{M}^{(0)}, \rho_{B}^{(0)}) = (7.1^\circ, -3.6^\circ)$, to avoid the complete flat state which hinders the mountain and valley crease assignments in the numerical calculations.
 Note that we maintain the mountain and valley crease assignments in the simulations.
 Also, we use the identical $\Delta \rho_C$ for every controlled unit cell to simplify the analysis.
 
 Figure~\ref{fig:MultiTrans}(a) shows the simulation results for the two different cases: (i) the first and second Miura-ori unit cells ($n=1,2$) are controlled, and (ii) first three unit cells ($n=1,2,3$) are controlled (see also Supplementary Movie S3 for the grasping animations).
Unlike the uniform grasping motion (see the inset illustrations in Fig.~\ref{fig:Geometry}(d)), the case where two unit cells ($n=1,2$) are controlled shows pinch-like folding motion of $n=1$ and 2 Miura-ori unit cells (analogous to a human hand pinching an object between thumb and index fingers).
Also, the leaf-out origami with three unit controlled ($n=1,2,3$) exhibits alligator-clip-like grasping motion, like holding an object between three tips of $n=1,2,3$ unit cells and two tips of the other two (see the inset (ii) in Fig.~\ref{fig:MultiTrans}(a)).
In addition, we examine the energy change 
during these two grasping motions, compared with the energy landscape for the uniform grasping motion.
Here, $\left( \bar{\rho}_M, \bar{\rho}_B \right) = (60^\circ, -120^\circ)$ is used.
In Fig.~\ref{fig:MultiTrans}(b), the energy changes during grasping are plotted as a function of the controlled angle ($\Delta \rho_C$), and the cases (i) and (ii) are denoted by the red and blue curves, respectively, as well as the uniform grasping case (grey curve).
Our energy analysis clearly indicates the energy minimum state along all grasping paths.

We further explore unique grasping motions by controlling different unit cells on the structure.
Figure~\ref{fig:MultiTrans}(c) shows 
three additional example motions 
and the three grasping motions discussed above (see Supplementary Movie S3 for each grasping motion).
In this figure, by introducing two variables calculated by $\rho_{M,2}-\rho_{M,4}$ and $\rho_{M,3}-\rho_{M,5}$, we plot the folding paths in the configuration space as a function of $\Delta \rho_C$ to characterize different configurations of the leaf-out origami.
Interestingly, we find various folded shapes from our numerical simulations, particularly two controlled unit cases exhibiting different folding paths and final configurations depending on selected two unit elements (see the red and magenta curves in Fig.~\ref{fig:MultiTrans}(c) for the two different cases of the leaf-out origami with $n=1,2$ and $n=1,3$ unit cells controlled).
Note that these folding motions obtained from the rigid origami model are an example of possible folding motions by searching a valid configuration at each iteration step in the numerical calculation.
Therefore, there are opportunities to achieve more diverse grasping modes in various ways, e.g., using different controlled angles for different unit cells, enabling the reversal of mountain/valley creases, etc., which can be useful for self-adaptive grasping for myriad target shapes.

\section*{Conclusion}

We have numerically and experimentally studied the unique snap-through behavior of leaf-out origami for grasping devices. 
We analyzed the multi-transformable nature of the leaf-out structure by employing a rigid origami model, and found that the leaf-out origami structure exhibits a tailorable bistable energy landscape, enabling controllable snap-through behavior by varying the rest angles $\bar{\rho}_j$. 
To demonstrate the adaptive grasping motion of the leaf-out origami, we have designed and fabricated multi-layered paper prototypes with customized hinges.
We have conducted drop tests by varying the drop height of a spherical target object, and our prototypes have exhibited controllable grasping motion through a triggering point that can be altered by the tailored energy landscape.
In addition, we have explored various grasping motions by selectively controlling the unit cell components of the leaf-out origami, and our numerical results have suggested the potential for highly adaptive grasping motions for different target shapes, inspired by human hands and other examples from nature.

In this study, we focused on showcasing the feasibility of grasping objects within a specific range of impact forces, but the leaf-out origami can also capture objects of varying masses and volumes that impact the structure off-center. Due to the scale-independent nature of rigid origami structures, a grasping device based on the leaf-out design can be fabricated not only for centimeter-scale target objects, but also for meter scale applications (e.g., robotic arm in space) and micron-/nano-scale operations (e.g., surgical operations and cleanings/examinations inside small diameter pipes. For example, see Supplementary Fig.~3 for the demonstration of the snap-through behavior of our millimeter-scale prototype (also see Supplementary Movie S4 for the grasping motion of the millimeter-scale prototype). 
This study elucidates the potential applications of the leaf-out origami structure as a capture mechanism within systems that require lightweight, easily tunable, and autonomous grasping. 

\section*{Methods}
\begin{figure}[htbp]
\centerline{ \includegraphics[width=0.65\textwidth]{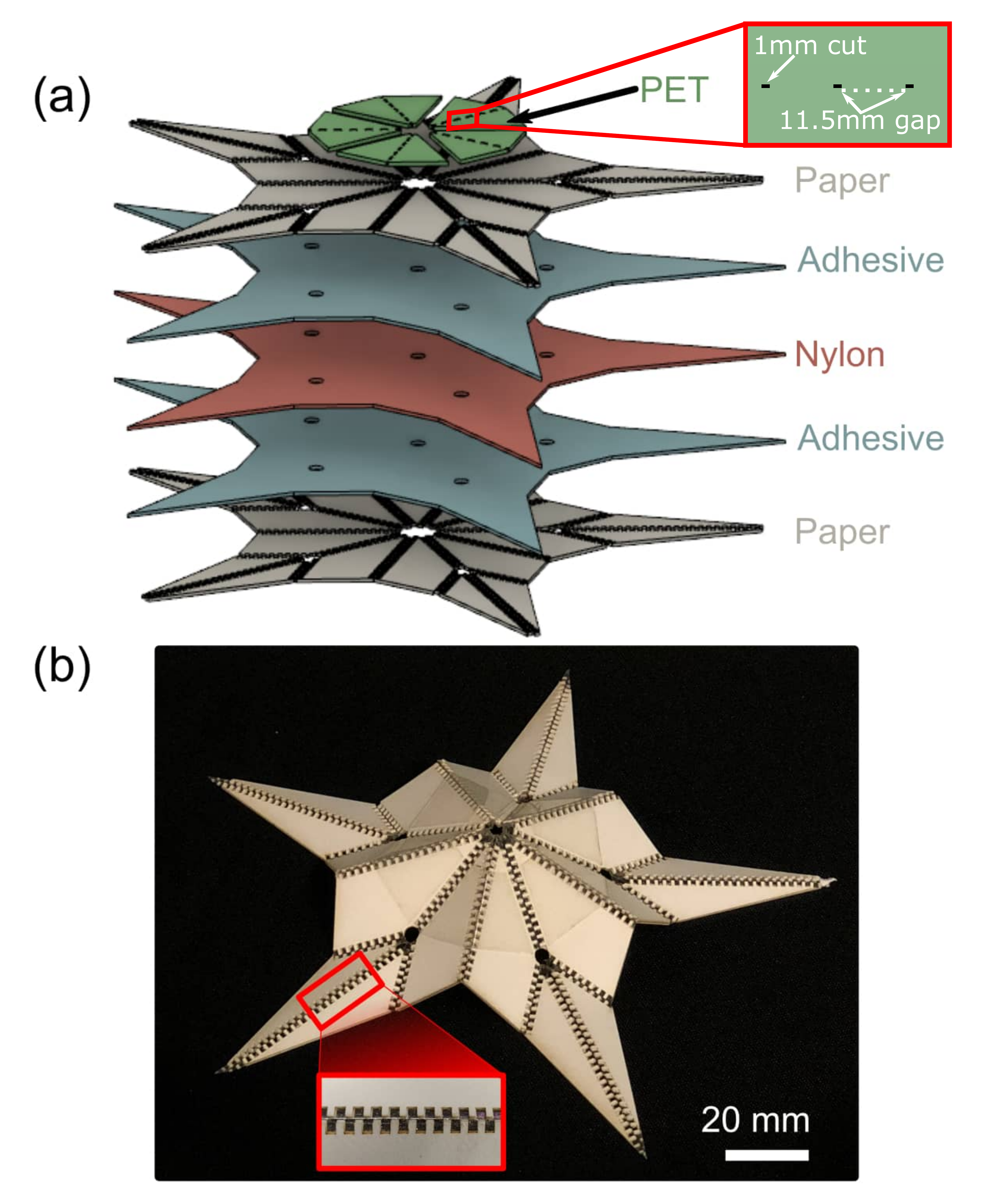}}
\caption{Paper leaf-out prototype fabrication by layer. (a) Layout of the composite structure consisting of paper, adhesive, Nylon, and PET layers. The layers also showcase the unique cut pattern required at each step of the process. (b) The photograph shows one of our handcrafted prototypes. The inset shows the magnified view of our comb-like crease design.}
\label{fig:Fabrication}
\end{figure}

\begin{figure}[htbp]
\centerline{ \includegraphics[width=1.\textwidth]{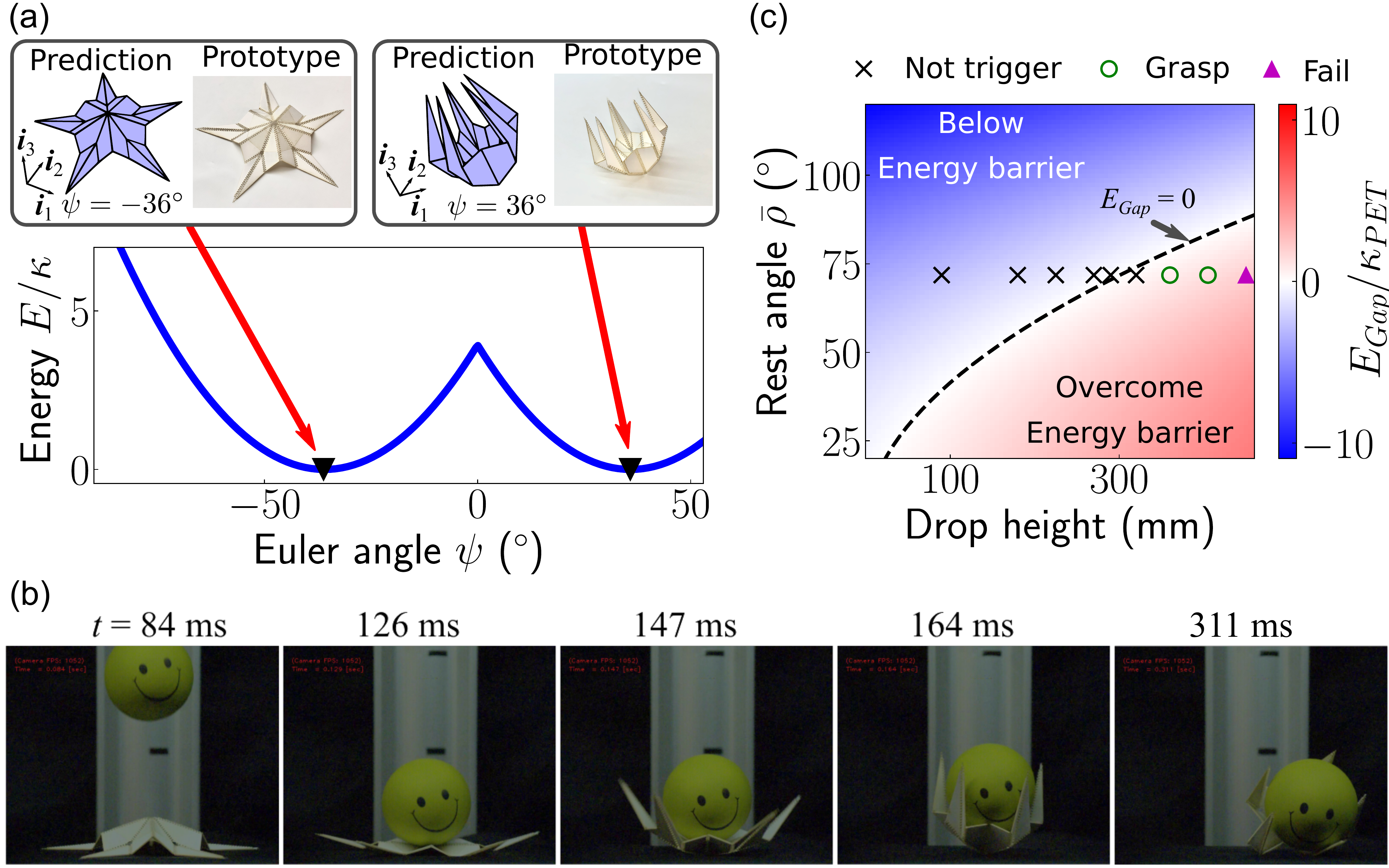}}
\caption{Experimental demonstration of leaf-out origami grasping motion. (a) Bistable energy landscape for our paper prototype with PET hinges. The left and right insets show the comparison between the predicted configuration and the physical prototype for open and closed states, respectively. (b) Snapshots of the grasping motion of the leaf-out prototype from the high speed camera images. The drop height is 360 mm. (c) We calculate the gap $E_{Gap}$ between the energy barrier to trigger snap-through behavior ($\Delta E_g$) and the potential energy $E_{ball}(h)$ where $h$ is the drop height. This energy gap is plotted as a function of the drop height ($h$) and the rest angle ($\bar{\rho}$) for the creases with PET. The dashed line indicates $E_{Gap}=0$. The cross, circle, triangle markers indicate the three different results of the drop test; (Cross markers) the grasping motion was not triggered, (Circle) the prototype grasped the ball, and (Triangle) the grasping motion was triggered, but the prototype failed to hold the ball.}
\label{fig:DropTest}
\end{figure}

\begin{figure}[htbp]
\centerline{ \includegraphics[width=1.\textwidth]{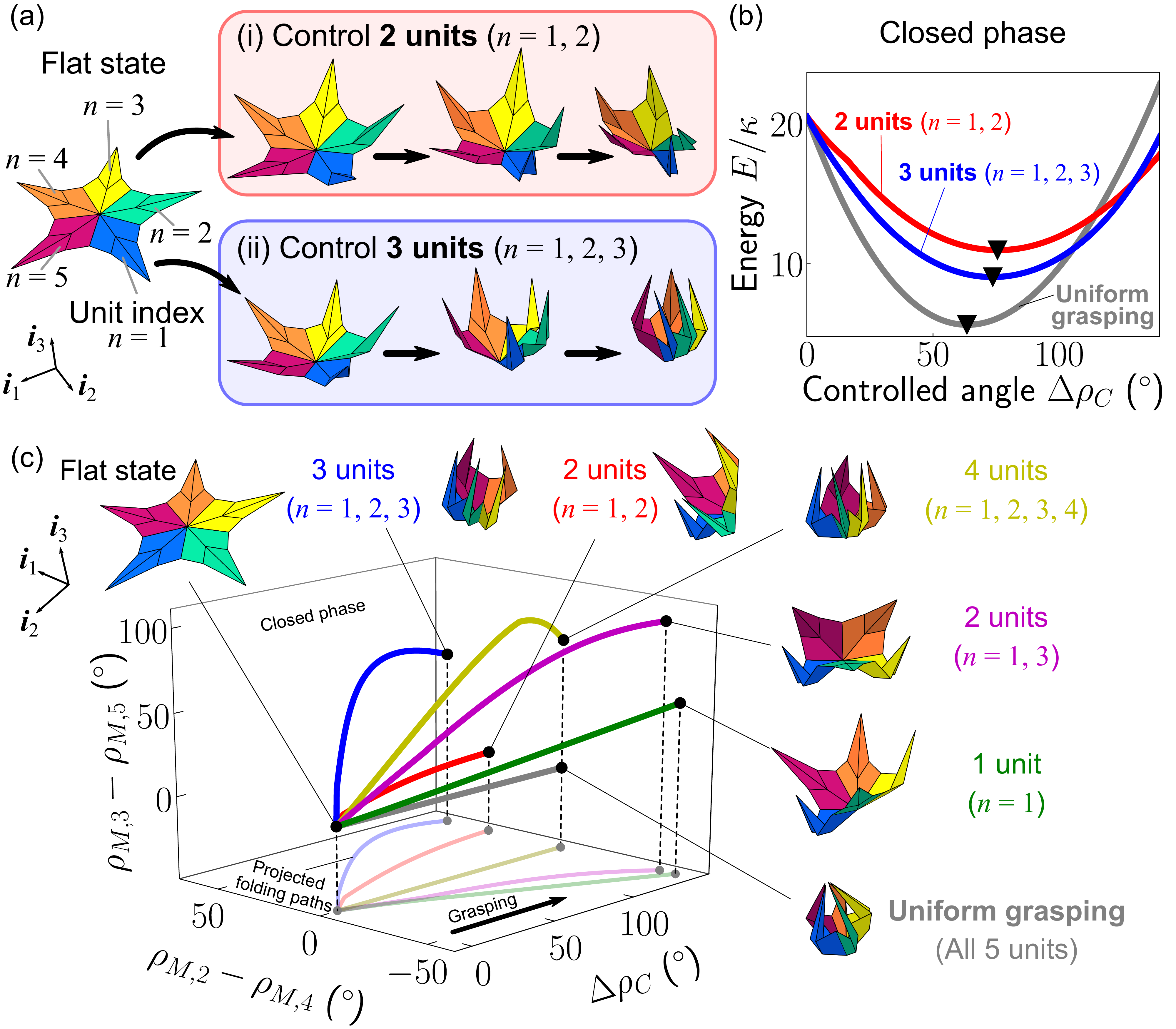}}
\caption{Multi-transformable grasping motions from the flat to closed states. (a) We explore different grapsing motions by controlling the folding angles of (i) the two unit cells ($n=1,2$, i.e., $\rho_{M,1}$ and $\rho_{M,2}$) and (ii) the three unit cells ($n=1,2,3$). (b) We calculate the energy landscapes for these two cases as well as the uniform grasping motion. The identical spring constant ($\kappa$) is used for all the crease lines, and $\left( \bar{\rho}_M, \bar{\rho}_B \right) = (60^\circ, -120^\circ)$ is used to calculate the energy curves. (c) Various different grasping motions are explored and plotted in the 3D configuration space. Each folding path is projected onto the $xy$-plane at the bottom.}
\label{fig:MultiTrans}
\end{figure}

\textbf{Fabrication of the paper prototype}
 To empirically test the leaf-out structure we utilized a CO$_2$ laser system to fabricate our centimeter-scale prototypes and a UV laser system to fabricate our millimeter-scale prototypes. We used construction paper (Strathmore 500 Series 3-PLY BRISTOL; thickness of the paper is 0.5 mm) and nylon (thickness of 0.025 mm) to maintain the flat surfaces on each unit cell, and polyester plastic (PET; thickness of 0.25 mm) to define the rigidity along the crease lines, as seen in Fig.~\ref{fig:Fabrication}(a). Between the bottom layer of construction paper and the nylon we used a 3M Very High Bond adhesive (VHB), and between the nylon and top layer of construction paper we used an Archival Double Tack Adhesive (ADTA). The ADTA's adhesive covering was left along each crease line during the top construction paper layer cut because these prototypes were fabricated by being stacked, so the covering protected the nylon layer from being pierced by the laser during the cut for the top construction paper layer. The ADTA was used for the top layer adhesive because it was thicker and therefore more protective, as well as being less adhesive and easier to remove after the final cut for the structure was made.

Each layer of our prototypes were constructed using either the laser cutting machine (VLS 4.6, Universal Laser Systems) or the diode-pumped solid-state frequency tripled Nd:Yag laser with 355nm wavelength (PhotoMachining, Inc.). The UV laser system was used for the millimeter-scale prototypes because the CO$_2$ laser system was not able to accurately cut at a millimeter scale. The crease lines of the leaf-out's paper layer were designed to
prevent torsional rotations along the creases without having an effect on the crease line's rigidity, as shown in Fig.~\ref{fig:Fabrication}(b). 
By using this crease pattern, without the PET layer, the structure's crease line stiffness is defined primarily by the flexible nylon layer. This increased the structure's resistance to fatigue, which improved the consistency of folding behavior and enhanced the repeatability of tests. Once the PET layer is applied, the crease line stiffness is assumed to be defined solely by the PET layer due to the pliability of nylon. We varied the rigidity of the boundary edges, see Fig.~\ref{fig:Geometry}(a), by altering the distance between 1.0 mm cuts in the PET layer as shown in Fig.~\ref{fig:Fabrication}(a). We found that tuning the stiffness of only the boundary edges, and leaving the other crease lines to be defined solely by the nylon layer, maximized the structure's ability to grasp onto dropped objects.

\textbf{Drop test}
To experimentally verify our analytically derived energy barrier, we fabricated a leaf-out structure with a PET layer crease line consisting of 1 mm cuts with 11.5 mm gaps in between each cut (see Fig.~\ref{fig:Fabrication}(a) for visualization). The object we dropped was a polyurethane ball of mass $m_{ball}=22.3$ g and volume $v_{ball}=179.6$ cm$^3$ ($R_{ball}=3.5$ cm).
We placed the leaf-out on top of an aluminum plate in front of the ball drop device, with drop height measured from the top of the aluminum plate to the bottom of the ball. The ball was pinched in place over the center of the leaf-out structure by pressure from a screw. To drop the ball with zero initial velocity we simply unscrewed the bolt to release the ball.

\section*{Data Availability}
Data supporting the findings of this study are available from the corresponding author on request.

\begin{acknowledgments}
We thank Professors Sawyer Fuller and Shyam Gollakota for their help with the fabrication of the millimeter scale Leafout prototype.
J.Y. and H.Y. are also grateful for the support from the 
U.S. National Science Foundation
(1553202 and 1933729) and the Washington Research Foundation.
H.Y. and J.R.R. gratefully acknowledge support from the Army Research Office award number W911NF-17–1–0147 and Air Force Office of Scientific Research award number FA9550-19-1-0285.
K.J. thanks the Department of Education's Ronald E. McNair Postbaccalaureate Achievement Program, the National Science Foundation's Research Experience for Undergraduates program and the Washington Research Foundation for their support. K.Y. is supported by the Funai Foundation for Information Technology.

\end{acknowledgments}

\section*{Author contributions}
H.Y. and K.J. proposed the research; H.Y. performed the modeling and numerical analysis; K.J., H.Y., V.A. and K.Y. conducted the experiments; J.Y. provided guidance throughout the research. H.Y., K.J., J.R.R and J.Y. prepared the manuscript.

\section*{Additional information}
The authors declare that they have no competing financial interests. Correspondence and requests for materials should be addressed to Jinkyu Yang~(email: jkyang@aa.washington.edu).

\bibliography{main.bib}

\begin{thebibliography}{20}
\expandafter\ifx\csname natexlab\endcsname\relax\def\natexlab#1{#1}\fi
\expandafter\ifx\csname bibnamefont\endcsname\relax
  \def\bibnamefont#1{#1}\fi
\expandafter\ifx\csname bibfnamefont\endcsname\relax
  \def\bibfnamefont#1{#1}\fi
\expandafter\ifx\csname citenamefont\endcsname\relax
  \def\citenamefont#1{#1}\fi
\expandafter\ifx\csname url\endcsname\relax
  \def\url#1{\texttt{#1}}\fi
\expandafter\ifx\csname urlprefix\endcsname\relax\def\urlprefix{URL }\fi
\providecommand{\bibinfo}[2]{#2}
\providecommand{\eprint}[2][]{\url{#2}}

\bibitem[{\citenamefont{Zang et~al.}(2020)\citenamefont{Zang, Liao, Lang, Zhao,
  Yuan, and Feng}}]{Zang2020}
\bibinfo{author}{\bibfnamefont{H.}~\bibnamefont{Zang}},
  \bibinfo{author}{\bibfnamefont{B.}~\bibnamefont{Liao}},
  \bibinfo{author}{\bibfnamefont{X.}~\bibnamefont{Lang}},
  \bibinfo{author}{\bibfnamefont{Z.-L.} \bibnamefont{Zhao}},
  \bibinfo{author}{\bibfnamefont{W.}~\bibnamefont{Yuan}}, \bibnamefont{and}
  \bibinfo{author}{\bibfnamefont{X.-Q.} \bibnamefont{Feng}},
  \bibinfo{journal}{Applied Physics Letters} \textbf{\bibinfo{volume}{116}},
  \bibinfo{pages}{023701} (\bibinfo{year}{2020}).

\bibitem[{\citenamefont{Guo et~al.}(2015)\citenamefont{Guo, Dai, Han, Xie,
  Chao, and Chen}}]{Guo2015}
\bibinfo{author}{\bibfnamefont{Q.}~\bibnamefont{Guo}},
  \bibinfo{author}{\bibfnamefont{E.}~\bibnamefont{Dai}},
  \bibinfo{author}{\bibfnamefont{X.}~\bibnamefont{Han}},
  \bibinfo{author}{\bibfnamefont{S.}~\bibnamefont{Xie}},
  \bibinfo{author}{\bibfnamefont{E.}~\bibnamefont{Chao}}, \bibnamefont{and}
  \bibinfo{author}{\bibfnamefont{Z.}~\bibnamefont{Chen}},
  \bibinfo{journal}{Journal of the Royal Society Interface}
  \textbf{\bibinfo{volume}{12}}, \bibinfo{pages}{20150598}
  (\bibinfo{year}{2015}).

\bibitem[{\citenamefont{Mu et~al.}(2018)\citenamefont{Mu, Wang, Yan, Li, Wang,
  Gao, Hou, Pham, Wu, Zhang et~al.}}]{Mu2018}
\bibinfo{author}{\bibfnamefont{J.}~\bibnamefont{Mu}},
  \bibinfo{author}{\bibfnamefont{G.}~\bibnamefont{Wang}},
  \bibinfo{author}{\bibfnamefont{H.}~\bibnamefont{Yan}},
  \bibinfo{author}{\bibfnamefont{H.}~\bibnamefont{Li}},
  \bibinfo{author}{\bibfnamefont{X.}~\bibnamefont{Wang}},
  \bibinfo{author}{\bibfnamefont{E.}~\bibnamefont{Gao}},
  \bibinfo{author}{\bibfnamefont{C.}~\bibnamefont{Hou}},
  \bibinfo{author}{\bibfnamefont{A.~T.~C.} \bibnamefont{Pham}},
  \bibinfo{author}{\bibfnamefont{L.}~\bibnamefont{Wu}},
  \bibinfo{author}{\bibfnamefont{Q.}~\bibnamefont{Zhang}},
  \bibnamefont{et~al.}, \bibinfo{journal}{Nature Communications}
  \textbf{\bibinfo{volume}{9}}, \bibinfo{pages}{590} (\bibinfo{year}{2018}).

\bibitem[{\citenamefont{Miura}(1985)}]{Miura1985}
\bibinfo{author}{\bibfnamefont{K.}~\bibnamefont{Miura}}, \bibinfo{journal}{The
  Institute of Space and Astronautical Science Report No.618} pp.
  \bibinfo{pages}{1--9} (\bibinfo{year}{1985}).

\bibitem[{\citenamefont{Kobayashi et~al.}(1998)\citenamefont{Kobayashi,
  Kresling, and Vincent}}]{Kobayashi1998}
\bibinfo{author}{\bibfnamefont{H.}~\bibnamefont{Kobayashi}},
  \bibinfo{author}{\bibfnamefont{B.}~\bibnamefont{Kresling}}, \bibnamefont{and}
  \bibinfo{author}{\bibfnamefont{J.~F.~V.} \bibnamefont{Vincent}},
  \bibinfo{journal}{Proceedings of the Royal Society B: Biological Sciences}
  \textbf{\bibinfo{volume}{265}}, \bibinfo{pages}{147} (\bibinfo{year}{1998}).

\bibitem[{\citenamefont{{De Focatiis} and Guest}(2002)}]{Focatiis2002}
\bibinfo{author}{\bibfnamefont{D.~S.~A.} \bibnamefont{{De Focatiis}}}
  \bibnamefont{and} \bibinfo{author}{\bibfnamefont{S.~D.} \bibnamefont{Guest}},
  \bibinfo{journal}{Philosophical transactions. Series A, Mathematical,
  physical, and engineering sciences} \textbf{\bibinfo{volume}{360}},
  \bibinfo{pages}{227} (\bibinfo{year}{2002}).

\bibitem[{\citenamefont{Mahadevan and Rica}(2005)}]{Mahadevan2005}
\bibinfo{author}{\bibfnamefont{L.}~\bibnamefont{Mahadevan}} \bibnamefont{and}
  \bibinfo{author}{\bibfnamefont{S.}~\bibnamefont{Rica}},
  \bibinfo{journal}{Science} \textbf{\bibinfo{volume}{307}},
  \bibinfo{pages}{1740} (\bibinfo{year}{2005}).

\bibitem[{\citenamefont{Faber et~al.}(2018)\citenamefont{Faber, Arrieta, and
  Studart}}]{Faber2018}
\bibinfo{author}{\bibfnamefont{J.~A.} \bibnamefont{Faber}},
  \bibinfo{author}{\bibfnamefont{A.~F.} \bibnamefont{Arrieta}},
  \bibnamefont{and} \bibinfo{author}{\bibfnamefont{A.~R.}
  \bibnamefont{Studart}}, \bibinfo{journal}{Science}
  \textbf{\bibinfo{volume}{359}}, \bibinfo{pages}{1386} (\bibinfo{year}{2018}).

\bibitem[{\citenamefont{Yasuda et~al.}(2016)\citenamefont{Yasuda, Chen, and
  Yang}}]{Yasuda2016}
\bibinfo{author}{\bibfnamefont{H.}~\bibnamefont{Yasuda}},
  \bibinfo{author}{\bibfnamefont{Z.}~\bibnamefont{Chen}}, \bibnamefont{and}
  \bibinfo{author}{\bibfnamefont{J.}~\bibnamefont{Yang}},
  \bibinfo{journal}{Journal of Mechanisms and Robotics}
  \textbf{\bibinfo{volume}{8}}, \bibinfo{pages}{031013} (\bibinfo{year}{2016}).

\bibitem[{\citenamefont{Hanna et~al.}(2014)\citenamefont{Hanna, Lund, Lang,
  Magleby, and Howell}}]{Hanna2014}
\bibinfo{author}{\bibfnamefont{B.~H.} \bibnamefont{Hanna}},
  \bibinfo{author}{\bibfnamefont{J.~M.} \bibnamefont{Lund}},
  \bibinfo{author}{\bibfnamefont{R.~J.} \bibnamefont{Lang}},
  \bibinfo{author}{\bibfnamefont{S.~P.} \bibnamefont{Magleby}},
  \bibnamefont{and} \bibinfo{author}{\bibfnamefont{L.~L.}
  \bibnamefont{Howell}}, \bibinfo{journal}{Smart Materials and Structures}
  \textbf{\bibinfo{volume}{23}}, \bibinfo{pages}{094009}
  (\bibinfo{year}{2014}).

\bibitem[{\citenamefont{Treml et~al.}(2018)\citenamefont{Treml, Gillman,
  Buskohl, and Vaia}}]{Treml2018}
\bibinfo{author}{\bibfnamefont{B.}~\bibnamefont{Treml}},
  \bibinfo{author}{\bibfnamefont{A.}~\bibnamefont{Gillman}},
  \bibinfo{author}{\bibfnamefont{P.}~\bibnamefont{Buskohl}}, \bibnamefont{and}
  \bibinfo{author}{\bibfnamefont{R.}~\bibnamefont{Vaia}},
  \bibinfo{journal}{Proceedings of the National Academy of Sciences}
  \textbf{\bibinfo{volume}{115}}, \bibinfo{pages}{6916} (\bibinfo{year}{2018}).

\bibitem[{\citenamefont{Waitukaitis et~al.}(2015)\citenamefont{Waitukaitis,
  Menaut, Chen, and van Hecke}}]{Waitukaitis2015}
\bibinfo{author}{\bibfnamefont{S.}~\bibnamefont{Waitukaitis}},
  \bibinfo{author}{\bibfnamefont{R.}~\bibnamefont{Menaut}},
  \bibinfo{author}{\bibfnamefont{B.~G.-g.} \bibnamefont{Chen}},
  \bibnamefont{and} \bibinfo{author}{\bibfnamefont{M.}~\bibnamefont{van
  Hecke}}, \bibinfo{journal}{Physical Review Letters}
  \textbf{\bibinfo{volume}{114}}, \bibinfo{pages}{055503}
  (\bibinfo{year}{2015}).

\bibitem[{\citenamefont{Tachi}(2009)}]{Tachi2009}
\bibinfo{author}{\bibfnamefont{T.}~\bibnamefont{Tachi}}, in
  \emph{\bibinfo{booktitle}{Origami 4}}, edited by
  \bibinfo{editor}{\bibfnamefont{R.~J.} \bibnamefont{Lang}}
  (\bibinfo{publisher}{A K Peters}, \bibinfo{address}{Natick, MA},
  \bibinfo{year}{2009}), pp. \bibinfo{pages}{175--187}.

\bibitem[{\citenamefont{Tachi}(2010)}]{Tachi2010}
\bibinfo{author}{\bibfnamefont{T.}~\bibnamefont{Tachi}}, in
  \emph{\bibinfo{booktitle}{Proceedings of the International Association for
  Shell and Spatial Structures (IASS) Symposium 2010, Shanghai}}
  (\bibinfo{address}{Shanghai, China}, \bibinfo{year}{2010}), pp.
  \bibinfo{pages}{771--782}.

\bibitem[{\citenamefont{Truby et~al.}(2018)\citenamefont{Truby, Wehner,
  Grosskopf, Vogt, Uzel, Wood, and Lewis}}]{Truby2018}
\bibinfo{author}{\bibfnamefont{R.~L.} \bibnamefont{Truby}},
  \bibinfo{author}{\bibfnamefont{M.}~\bibnamefont{Wehner}},
  \bibinfo{author}{\bibfnamefont{A.~K.} \bibnamefont{Grosskopf}},
  \bibinfo{author}{\bibfnamefont{D.~M.} \bibnamefont{Vogt}},
  \bibinfo{author}{\bibfnamefont{S.~G.} \bibnamefont{Uzel}},
  \bibinfo{author}{\bibfnamefont{R.~J.} \bibnamefont{Wood}}, \bibnamefont{and}
  \bibinfo{author}{\bibfnamefont{J.~A.} \bibnamefont{Lewis}},
  \bibinfo{journal}{Advanced Materials} \textbf{\bibinfo{volume}{30}},
  \bibinfo{pages}{1706383} (\bibinfo{year}{2018}).

\bibitem[{\citenamefont{Edmondson et~al.}(2013)\citenamefont{Edmondson, Bowen,
  Grames, Magleby, Howell, and Bateman}}]{Edmondson2013}
\bibinfo{author}{\bibfnamefont{B.~J.} \bibnamefont{Edmondson}},
  \bibinfo{author}{\bibfnamefont{L.~A.} \bibnamefont{Bowen}},
  \bibinfo{author}{\bibfnamefont{C.~L.} \bibnamefont{Grames}},
  \bibinfo{author}{\bibfnamefont{S.~P.} \bibnamefont{Magleby}},
  \bibinfo{author}{\bibfnamefont{L.~L.} \bibnamefont{Howell}},
  \bibnamefont{and} \bibinfo{author}{\bibfnamefont{T.~C.}
  \bibnamefont{Bateman}}, in \emph{\bibinfo{booktitle}{Proceedings of the ASME
  2013 Conference on Smart Materials, Adaptive Structures and Intelligent
  Systems}} (\bibinfo{year}{2013}), p. \bibinfo{pages}{V001T01A027}.

\bibitem[{\citenamefont{Mintchev et~al.}(2018)\citenamefont{Mintchev, Shintake,
  and Floreano}}]{Mintchev2018}
\bibinfo{author}{\bibfnamefont{S.}~\bibnamefont{Mintchev}},
  \bibinfo{author}{\bibfnamefont{J.}~\bibnamefont{Shintake}}, \bibnamefont{and}
  \bibinfo{author}{\bibfnamefont{D.}~\bibnamefont{Floreano}},
  \bibinfo{journal}{Science Robotics} \textbf{\bibinfo{volume}{3}},
  \bibinfo{pages}{eaau0275} (\bibinfo{year}{2018}).

\bibitem[{\citenamefont{Li et~al.}(2019)\citenamefont{Li, Stampfli, Xu, Malkin,
  Diaz, Rus, and Wood}}]{Li2019}
\bibinfo{author}{\bibfnamefont{S.}~\bibnamefont{Li}},
  \bibinfo{author}{\bibfnamefont{J.~J.} \bibnamefont{Stampfli}},
  \bibinfo{author}{\bibfnamefont{H.~J.} \bibnamefont{Xu}},
  \bibinfo{author}{\bibfnamefont{E.}~\bibnamefont{Malkin}},
  \bibinfo{author}{\bibfnamefont{E.~V.} \bibnamefont{Diaz}},
  \bibinfo{author}{\bibfnamefont{D.}~\bibnamefont{Rus}}, \bibnamefont{and}
  \bibinfo{author}{\bibfnamefont{R.~J.} \bibnamefont{Wood}}, in
  \emph{\bibinfo{booktitle}{Proceedings - IEEE International Conference on
  Robotics and Automation}} (\bibinfo{year}{2019}), vol.
  \bibinfo{volume}{2019-May}, pp. \bibinfo{pages}{7401--7408}.

\bibitem[{\citenamefont{Hu et~al.}(2020)\citenamefont{Hu, Wang, Cheng, and
  Bao}}]{Hu2020}
\bibinfo{author}{\bibfnamefont{F.}~\bibnamefont{Hu}},
  \bibinfo{author}{\bibfnamefont{W.}~\bibnamefont{Wang}},
  \bibinfo{author}{\bibfnamefont{J.}~\bibnamefont{Cheng}}, \bibnamefont{and}
  \bibinfo{author}{\bibfnamefont{Y.}~\bibnamefont{Bao}},
  \bibinfo{journal}{Science Progress} \textbf{\bibinfo{volume}{103}},
  \bibinfo{pages}{0036850420946162} (\bibinfo{year}{2020}).

\bibitem[{\citenamefont{Yasuda et~al.}(2019)\citenamefont{Yasuda,
  Gopalarethinam, Kunimine, Tachi, and Yang}}]{Yasuda2019tmp}
\bibinfo{author}{\bibfnamefont{H.}~\bibnamefont{Yasuda}},
  \bibinfo{author}{\bibfnamefont{B.}~\bibnamefont{Gopalarethinam}},
  \bibinfo{author}{\bibfnamefont{T.}~\bibnamefont{Kunimine}},
  \bibinfo{author}{\bibfnamefont{T.}~\bibnamefont{Tachi}}, \bibnamefont{and}
  \bibinfo{author}{\bibfnamefont{J.}~\bibnamefont{Yang}},
  \bibinfo{journal}{Advanced Engineering Materials}
  \textbf{\bibinfo{volume}{21}}, \bibinfo{pages}{1900562}
  (\bibinfo{year}{2019}).

\end{thebibliography}

\newpage

\end{document}